%% file: main.tex
\documentclass[10pt,twocolumn,letterpaper]{article}

\usepackage{iccv}              


\usepackage{multirow}

\definecolor{iccvblue}{rgb}{0.21,0.49,0.74}
\usepackage[pagebackref,breaklinks,colorlinks,allcolors=iccvblue]{hyperref}

\def\paperID{13920} 
\def\confName{ICCV}
\def\confYear{2025}

\title{Integrating Chain-of-Thought for Multimodal Alignment: A Study on \\ 3D Vision-Language Learning}

\author{
Yanjun Chen\textsuperscript{1,2}\thanks{Equal Contribution.},  
Yirong Sun\textsuperscript{2}\footnotemark[1],  
Xinghao Chen\textsuperscript{1,2},  
Jian Wang\textsuperscript{1},  
Xiaoyu Shen\textsuperscript{2},  
Wenjie Li\textsuperscript{1}\thanks{Corresponding Author.},  
Wei Zhang\textsuperscript{2}\thanks{Corresponding Author.} \\
\textsuperscript{1}Department of Computing, The Hong Kong Polytechnic University, Hong Kong, China \\
\textsuperscript{2}Digital Twin Institute, Eastern Institute of Technology, Ningbo, China \\
{\tt\small yan-jun.chen@connect.polyu.hk, zhw@eitech.edu.cn}
}

\begin{document}
\maketitle
\input{sec/0_abstract}    
\input{sec/1_intro}
\input{sec/2_relatedwork}
\input{sec/3_method}

\input{sec/4_experiments}
\input{sec/5_conclusion}
\newpage
{
    \small
    \bibliographystyle{ieeenat_fullname}
    \bibliography{main}
}

\input{sec/supplement}
\end{document}

%% file: sec/0_abstract.tex

\begin{abstract}
Chain-of-Thought (CoT) reasoning has proven effective in natural language tasks but remains underexplored in multimodal alignment. This study investigates its integration into 3D vision-language learning by embedding structured reasoning into alignment training. We introduce the \textit{3D-CoT Benchmark}, a dataset with hierarchical CoT annotations covering shape recognition, functional inference, and causal reasoning. Through controlled experiments, we compare CoT-structured and standard textual annotations across large reasoning models (LRMs) and large language models (LLMs). Our evaluation employs a dual-layer framework assessing both intermediate reasoning and final inference quality. Extensive experiments demonstrate that CoT significantly improves 3D semantic grounding, with LRMs leveraging CoT more effectively than LLMs. Furthermore, we highlight that annotation structure influences performance—explicit reasoning markers aid LLMs, while unmarked CoT better aligns with LRM inference patterns. Our analyses suggest that CoT is crucial for enhancing multimodal reasoning, with implications beyond 3D tasks.
\end{abstract}

%% file: sec/1_intro.tex
\section{Introduction}
\label{sec:intro}

Multimodal learning, which integrates diverse data modalities such as vision, audio, and text, has made significant progress in recent years. However, achieving robust alignment across modalities remains a fundamental challenge, particularly in 3D vision-language tasks. Compared to 2D images, 3D data introduce greater structural complexity, functional diversity, and interaction dependencies, making it more difficult for models to reason about object affordances, causal relationships, and scene dynamics.\footnote{Baseline comparisons are still in progress. Additional results will be updated in future versions of this manuscript.} \footnote{The dataset will be publicly available at: \href{https://huggingface.co/datasets/Battam/3D-CoT}{https://huggingface.co/datasets/Battam/3D-CoT}.}

\begin{figure}[ht!]
    \centering
    \includegraphics[width=0.45\textwidth]{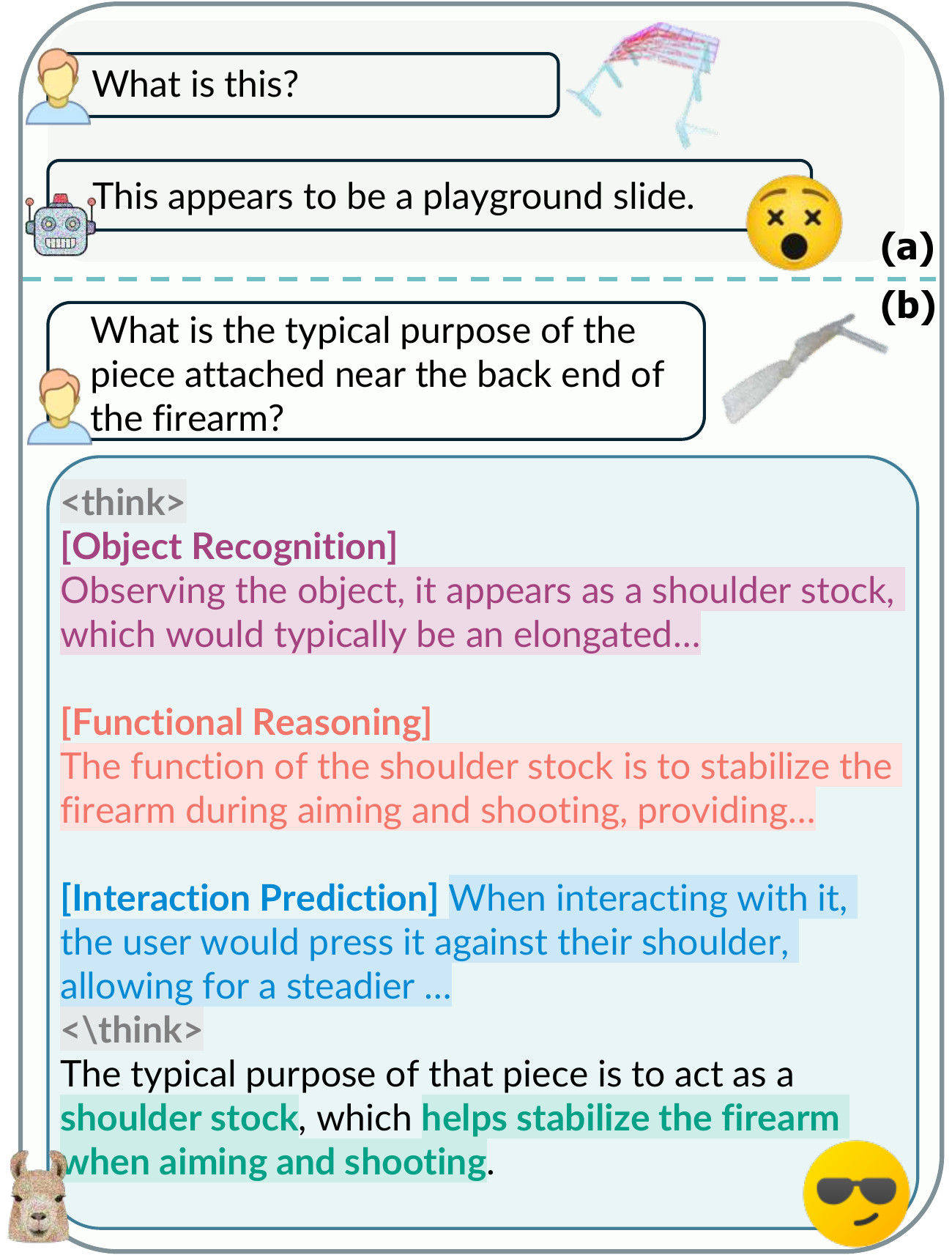}
    \caption{Illustration of 3D vision-language reasoning. (a) Traditional 3D vision-language alignment relies on static descriptions, overlooking hierarchical reasoning. (b) Our CoT-based approach encodes intermediate reasoning steps, enhancing semantic grounding and functional inference.}
    \label{fig:overview}
\end{figure}

Recent multimodal large language models (MLLMs), such as ShapeLLM~\cite{qi2024shapellm} and ULIP-2~\cite{xue2024ulip}, have demonstrated strong text-to-3D alignment capabilities, enabling applications like zero-shot classification and retrieval. However, these models predominantly rely on static textual descriptions and do not explicitly model structured reasoning. As a result, they struggle to capture the hierarchical relationships and causal mechanisms that underlie 3D interactions, limiting their ability to reason about object affordances and functional dependencies.

Chain-of-Thought (CoT) reasoning has been shown to enhance complex problem-solving in natural language processing by structuring tasks into intermediate, interpretable steps~\cite{wei2022chain}. Despite its success in textual domains, its role in multimodal learning remains underexplored. We hypothesize that incorporating CoT into 3D vision-language models can improve semantic grounding by explicitly linking object geometry to its affordances and interactions.  
As illustrated in Figure~\ref{fig:overview}, CoT-based annotations are able to improve semantic understanding and functional reasoning in 3D vision-language alignment.
This approach could enable models to reason beyond surface-level descriptions and develop a deeper understanding of 3D environments.

With this in mind, we propose integrating CoT for 3D vision-language alignment. Specifically, we introduce the \textbf{3D-CoT Benchmark}, which augments existing 3D datasets~\cite{deitke2023objaverse, chang2015shapenet, collins2022abo, fu20213d} with hierarchical reasoning annotations covering shape recognition, functional inference, and causal reasoning. We conduct controlled experiments comparing standard textual annotations with CoT-structured annotations to assess their impact on different model architectures, including large reasoning models (LRMs) optimized for structured inference and general-purpose large language models (LLMs).

To rigorously evaluate model performance, we employ a \textbf{dual-layer evaluation framework} that separately quantifies intermediate reasoning quality and final inference correctness. We measure reasoning decomposition along three dimensions: object recognition, functional reasoning, and interaction prediction, each rated on a 1–5 scale. 
Additionally, we assess the factual consistency and coherence of final conclusions using truthfulness and completeness metrics. This evaluation framework allows us to isolate the contributions of CoT-enhanced training from broader multimodal learning improvements.
Experimental results demonstrate that CoT-structured annotations significantly improve 3D reasoning. Models trained with CoT exhibit enhanced alignment between textual descriptions and 3D structures, particularly in affordance recognition and interaction prediction. Moreover, LRMs leverage CoT more effectively than LLMs, suggesting that models designed for structured inference benefit more from reasoning-rich supervision. Annotation structure also plays a crucial role—explicit reasoning markers improve alignment in LLMs, while unmarked CoT annotations better align with LRM’s internal inference mechanisms. These findings suggest that CoT provides a fundamental bridge between 3D representations and high-level semantic reasoning.

In summary, our contributions are threefold: 1) We propose a novel CoT-driven framework specifically designed for 3D vision-language alignment. 2) We introduce the 3D-CoT Benchmark, a comprehensive dataset enriched with structured reasoning annotations to facilitate advanced multimodal learning. 3) We develop a dual-layer evaluation framework that systematically assesses the impact of CoT-enhanced training on model performance. Our findings underscore the broader applicability of CoT in multimodal learning, suggesting potential extensions to domains beyond 3D vision-language tasks. This work not only advances the field of 3D vision-language alignment but also opens new avenues for leveraging CoT in diverse multimodal contexts.



%% file: sec/2_relatedwork.tex
\section{Related Work}
\label{sec:related}

\subsection{Multimodal Alignment}

\paragraph{3D-Text Contrastive Pretraining}
In the 2D domain, large-scale methods such as CLIP~\cite{radford2021learning} and BLIP-2~\cite{li2023blip} excel at image-text alignment through contrastive objectives or learnable modules. Extending these advances to 3D often involves projecting point clouds into multi-view images, as in PointCLIP~\cite{zeng2023clip2}, which leverages 2D CLIP embeddings but risks losing spatial details critical for functional or part-level reasoning. To address this, ULIP~\cite{xue2023ulip} aligns point clouds, images, and text via tri-modal contrastive objectives and synthesized shape-image-text triplets on ShapeNet, enabling strong zero-shot performance on tasks like ModelNet40 classification and ScanObjectNN recognition. Subsequent methods scale up this idea: ULIP-2~\cite{xue2024ulip} enriches shape descriptions via GPT-based captioning on more diverse 3D-text pairs, and OpenShape~\cite{Liu_2023_NeurIPS} trains across unified 3D datasets with over 1,156 categories. These larger vocabularies and broader datasets push zero-shot accuracies to new levels (e.g., exceeding 46\% top-1 accuracy on Objaverse-LVIS), underscoring the growing feasibility of \emph{open-world} 3D understanding.

\paragraph{3D Scene Understanding and Rich Text Supervision}
Beyond object-centric tasks, 3D scene-level alignment introduces added complexity in modeling spatial arrangements. 3D-VisTA~\cite{Zhu_2023_ICCV} tackles entire 3D scans by jointly encoding scene geometry and textual descriptions via a transformer-based pipeline, pretraining on 3K scans and 278K GPT-generated captions (ScanScribe). This approach benefits visual grounding and question answering without specialized architectures. Alternatively, synthetic large-scale corpora offer richer supervision. SynVL3D~\cite{Yang_2024_arXiv} constructs 10K indoor scenes paired with 1M captions covering object, view, and room levels. This synthetic data drives robust generalization to real scans, highlighting the value of broad, precisely annotated 3D-text pairs. As scene complexity grows, such strategies become vital for capturing nuanced spatial relationships and object interactions.

\paragraph{Functional Semantics and Affordances}
Recent trends emphasize \emph{why} shapes matter and \emph{how} they can be used, rather than merely \emph{what} they are. Datasets like SceneFun3D~\cite{delitzas2024scenefun3d} annotate task-specific parts (e.g., knobs, handles) to link language instructions with functional geometry. Models focusing on open-vocabulary affordance detection (e.g., 3D-AffordanceLLM~\cite{chu20253daffordancellmharnessinglargelanguage}) interpret instructions to identify relevant surfaces or parts in point clouds. By inferring which actions an object supports, these systems move beyond identity-driven classification, leveraging detailed textual or symbolic cues to capture \emph{function}---a key aspect of 3D comprehension. As large-scale 3D-text pretraining increasingly incorporates part-level and functional descriptors, future methods stand poised to unlock deeper reasoning about the relationship between shape geometry and real-world usage.

\subsection{Chain-of-Thought Reasoning}

CoT prompting~\cite{wei2022chain, kojima2022large} guides LLMs to articulate step-by-step reasoning, yielding improvements in interpretability and logical consistency on tasks like arithmetic and commonsense QA. Self-consistency~\cite{wang2022self} further samples multiple reasoning paths, aggregating final predictions for better robustness.

\paragraph{CoT in Multimodal Tasks}
While initially developed for text-centric tasks, CoT prompting has been extended to image-text scenarios, where intermediate visual context shapes model reasoning. Multimodal-CoT~\cite{zhang2023multimodal} and KAM-CoT~\cite{Mondal_2024_AAAI} combine stepwise rationales with knowledge graphs or domain-specific information, achieving state-of-the-art performance on complex benchmarks like ScienceQA. In robotics, methods such as PaLM-E~\cite{driess2023palmeembodiedmultimodallanguage} and EmbodiedGPT~\cite{mu2023embodiedgpt} integrate CoT with egocentric visual inputs, enabling multi-step action planning and spatiotemporal decision-making. This paradigm underscores how explicit reasoning steps can mediate between raw perceptual features and structured plans.

\paragraph{3D-Specific Reasoning Scenarios}
Recent works have begun to incorporate CoT prompting into 3D tasks. 3D-LLM~\cite{Hong_2023_NeurIPS} fuses point-cloud encoders with LLMs to handle 3D QA, captioning, and sequential planning; it supplies multi-view image features to guide the chain-of-thought, allowing the model to reason about object arrangement and spatial cues. Further integrating simulation-based CoT, 3D-VLA~\cite{Zhen_2024_arXiv} iteratively imagines future scene states through diffusion models before planning, embedding a “what-if” loop in the CoT pipeline. Such approaches highlight CoT's potential for resolving 3D challenges like occlusion, geometry, and functional dependencies, suggesting explicit reasoning can improve both performance and interpretability in 3D domains.

\subsection{CoT for Multimodal Alignment}

Despite extensive progress in large-scale 3D vision-language alignment~\cite{xue2023ulip, xue2024ulip, Liu_2023_NeurIPS, Zhu_2023_ICCV} and promising results for CoT in other multimodal settings, their intersection remains nascent. Early evidence suggests that CoT can provide structured rationales vital for nuanced 3D understanding, from part-level semantics to scene-level affordances.

Most 3D alignment pipelines produce final outputs in a single pass, a CoT layer could systematically interpret geometric cues and textual context. Methods like TriCoLo~\cite{ruan2024tricolo} or OpenScene~\cite{peng2023openscene} might benefit from stepwise dissection of shape parts, object adjacencies, or functional regions, enabling deeper semantic alignment than one-shot embedding.
Because geometry in 3D directly correlates with object utility, CoT can articulate the \emph{why} behind an affordance label. For instance, 3D-AffordanceLLM~\cite{chu20253daffordancellmharnessinglargelanguage} links shape features to actionable insights (“this has a handle, so it can be grasped”), improving functional detection and part-level segmentation. Explicit reasoning sequences can clarify subtle distinctions among objects with similar shapes but different intended uses.
A future frontier is merging CoT with symbolic or causal reasoning. Neuro-symbolic 3D approaches~\cite{hsu2023ns3d} typically parse language into formal commands executed on structured scene graphs. Embedding a CoT mechanism here would let a model validate partial inferences by executing them in a physics engine or symbolic planner—akin to ``reasoning-by-execution'' frameworks in 2D~\cite{suris2023vipergpt}.

\paragraph{Benchmarking}
Emerging benchmarks increasingly require robust interpretability and logical coherence. Datasets like ScanQA~\cite{azuma2022scanqa} and Objaverse-LVIS~\cite{Liu_2023_NeurIPS} demand open-ended or fine-grained 3D understanding, while habitat-based QA tasks~\cite{Ghaffari_2024_EMNLP} and synthetic 3D corpora~\cite{Yang_2024_arXiv} stress causal and spatial reasoning. Incorporating CoT into evaluation protocols can pinpoint failure cases in geometric inference and highlight the synergy between structured reasoning and perceptual alignment. Recent progress in knowledge-grounded CoT~\cite{Mondal_2024_AAAI} hints that even smaller models, when equipped with coherent reasoning chains, can outperform larger LLMs in specialized domains.

Overall, the nascent fusion of \emph{3D vision-language pretraining} and \emph{chain-of-thought reasoning} signals a path toward higher-level 3D semantic understanding and more robust, interpretable performance. By unifying these two paradigms, future systems may handle not only object identity but also part-level geometry, affordances, and causal dependencies, all while articulating clear intermediate rationales for complex 3D tasks.

%% file: sec/3_method.tex
\section{Method}
\label{sec:method}

\begin{figure*}[t!]
    \centering
    \includegraphics[width=1.0\textwidth]{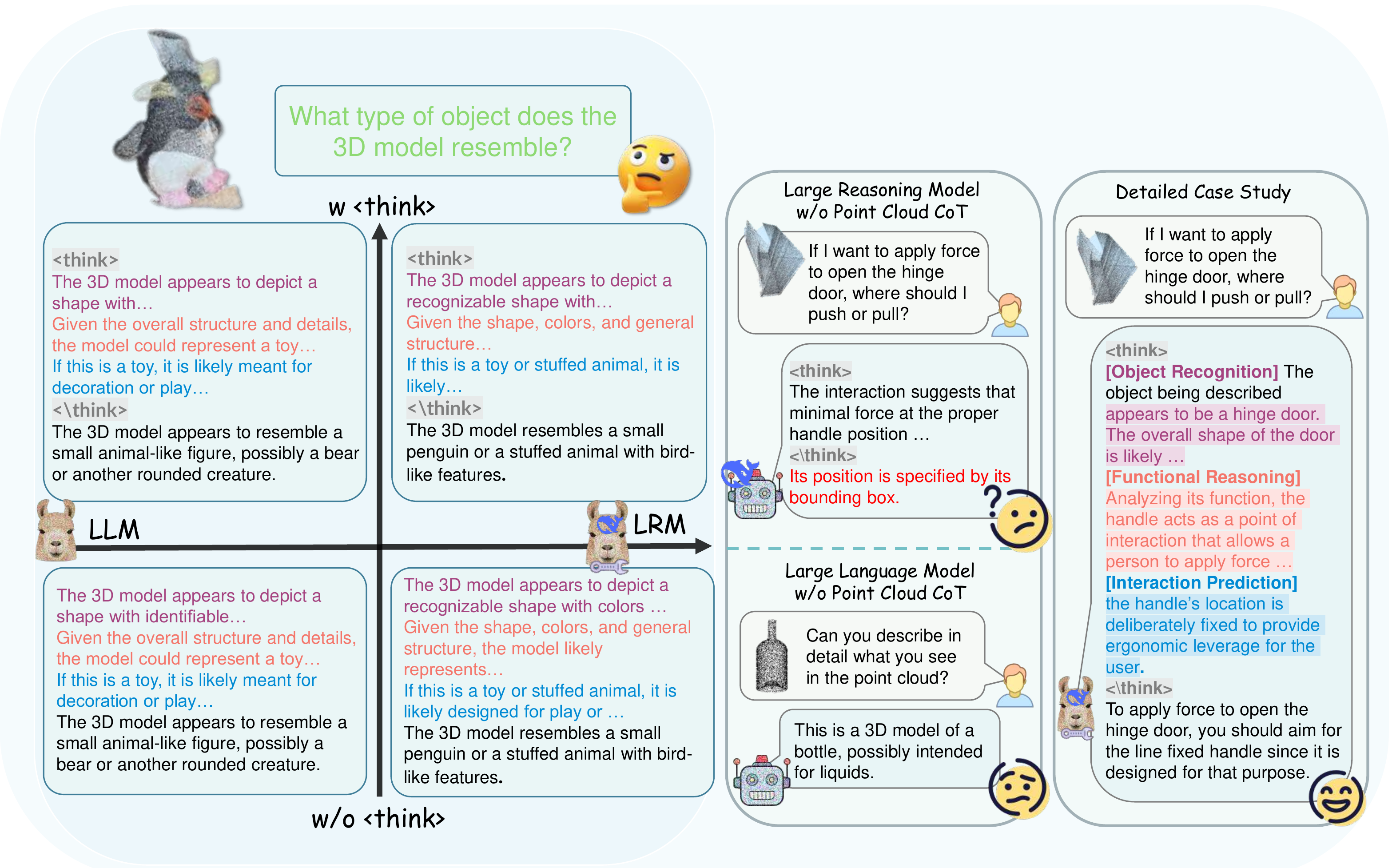}
    \caption{Overview of our Chain-of-Thought (CoT) reasoning for 3D vision-language alignment.  
    \textbf{Left}: Model behavior categorized by annotation type (Tagged vs. Unmarked) and model type (LLM vs. LRM).
    \textbf{Right}: Illustration of CoT's role in enhancing reasoning depth: Tagged CoT explicitly delineates steps, while Unmarked CoT fosters implicit integration. Overall, LLMs benefit from explicit segmentation, whereas LRMs align better with unmarked reasoning.}
    \label{fig:cot_types}
\end{figure*}

We propose a framework that integrates CoT reasoning into 3D vision-language alignment, leveraging structured annotations to enhance semantic grounding and functional inference. Our approach consists of four key components: an overview of the CoT integration pipeline (Section~\ref{subsec:method_overview}), the construction of the \textit{3D-CoT Benchmark} (Section~\ref{subsec:3dcot_benchmark}), dataset statistics highlighting the scale and composition of training and evaluation sets, and a multimodal alignment strategy that optimally fuses 3D representations with stepwise textual reasoning (Section~\ref{subsec:architecture}).

\subsection{Method Overview}
\label{subsec:method_overview}

We enhance 3D vision-language alignment by integrating structured reasoning into model training. As shown in Figure\,\ref{fig:overview}, our method encodes raw 3D shapes into a text-compatible embedding space and fuses them with CoT-structured textual cues to improve geometric understanding, functional inference, and causal reasoning. Given a 3D point set \(\mathbf{P}\) and a textual description \(\mathbf{T}\), we map \(\mathbf{P}\) to a latent vector \(\mathbf{z}_\text{3D}\) and \(\mathbf{T}\) to \(\mathbf{z}_\text{text}\), aligning them to facilitate multimodal reasoning beyond surface-level descriptions.

\subsection{3D-CoT Benchmark}
\label{subsec:3dcot_benchmark}

We aim to extend existing 3D datasets with structured reasoning annotations, facilitating a systematic study of CoT's impact on vision-language alignment. Each shape instance is annotated with:
\begin{itemize}[leftmargin=1.5em]
    \item \textbf{Object recognition}: Identifies object categories.
    \item \textbf{Functional inference}: Describes affordances and design features.
    \item \textbf{Causal reasoning}: Explains how structural attributes influence real-world interactions.
\end{itemize}
Our annotations are formatted in three styles:
(\emph{i})~\textbf{Tagged CoT}, which explicitly segments reasoning steps using \texttt{<think>} markers;  
(\emph{ii})~\textbf{Unmarked CoT}, which integrates multi-step reasoning into continuous text;  
(\emph{iii})~\textbf{No CoT}, which merely provides concise descriptions without structured reasoning.
In this manner, we construct a 3D-CoT Benchmark comprising a large-scale alignment dataset and two fine-tuning subsets, each targeting different aspects of CoT integration. The details are introduced as follows.


\paragraph{Data Sources.}  
We construct the benchmark using diverse 3D datasets, including Objaverse~\cite{deitke2023objaverse}, ShapeNet~\cite{chang2015shapenet}, ABO~\cite{collins2022abo}, and 3D-FUTURE~\cite{fu20213d}. These datasets offer extensive geometric and semantic annotations across various object categories.

\paragraph{CoT Annotation Pipeline.}  
We employ a two-stage pipeline to generate structured reasoning annotations, aligned with the reasoning hierarchy outlined in Section~\ref{subsec:3dcot_benchmark}.

\begin{enumerate}[leftmargin=1.2em]
    \item \textbf{LLM-Assisted Annotation:} GPT-4o~\cite{openai2024gpt4o} generates textual descriptions based on object recognition, functional inference, and causal reasoning. Each annotation follows one of the three CoT formats: Tagged, Unmarked, or No CoT.
    \item \textbf{Human Verification:} Experts manually review and refine \textbf{20\%} of the generated samples, ensuring factual accuracy, logical consistency, and completeness.
\end{enumerate}

\paragraph{Dataset Structure.}
\begin{itemize}[leftmargin=1.5em]
    \item \textbf{CoT-CAP3D (Alignment Dataset)}:  
    A collection of 1.51M shape-text pairs, each provided in two versions: with and without explicit \texttt{<think>} markers.
    \item \textbf{Fine-tuning Subsets}:  
    Focused datasets for targeted CoT training:
    \begin{itemize}[leftmargin=1.2em]
        \item \textbf{CoT-CAP3D-SFT}: 90K samples for shape recognition and basic affordance reasoning.
        \item \textbf{CoT-GApartNet-SFT}: 54K samples emphasizing part-level functionality and causal reasoning.
    \end{itemize}
    \item \textbf{Evaluation Sets}:  
    Each subset includes 100 test samples, enabling controlled comparisons between CoT and non-CoT settings.
\end{itemize}

\paragraph{Key Dataset Statistics.}
\begin{itemize}[leftmargin=1.5em]
    \item \textbf{CoT-CAP3D (Alignment Dataset)}:  
    1.51M shape-text pairs in two formats: with and without \texttt{<think>} markers. The only difference between these variants is the explicit segmentation of reasoning steps.
    
    \item \textbf{Fine-Tuning Subsets}:  
    Two supervised fine-tuning datasets for targeted CoT training:
    \begin{itemize}[leftmargin=1.2em]
        \item \textbf{CoT-CAP3D-SFT}: 90K samples covering shape recognition and basic affordance reasoning.
        \item \textbf{CoT-GApartNet-SFT}: 54K samples focusing on causal interactions and part-level affordances.
    \end{itemize}
    
    \item \textbf{Total Dataset Size}:  
    1.64M samples, incorporating both CoT and non-CoT data for comparative analysis.
    
    \item \textbf{Benchmark Evaluation Sets}:  
    100 test samples per subset, ensuring a controlled comparison across CoT-structured and non-CoT variants, including CoT-CAP3D, CoT-GApartNet, and their respective baselines.
\end{itemize}

\noindent
By integrating large-scale CoT annotations and unlabeled data, this benchmark systematically evaluates the impact of explicit reasoning markers and multi-step textual descriptions on semantic grounding in 3D vision-language tasks.

\subsection{3D Multimodal Alignment}
\label{subsec:architecture}

\paragraph{3D Encoder.}
We adopt the ReCon++ point cloud encoder~\cite{qi2024shapellm} to extract geometric descriptors from raw 3D data. The encoder employs farthest point sampling to select keypoints and aggregates local features, producing a global embedding \(\mathbf{e}_\text{3D} \in \mathbb{R}^{d}\). This embedding captures fine-grained structural details and overall contextual information, supporting object recognition and higher-level reasoning.

\paragraph{Projection Module.}
To align 3D and textual representations, we employ a projection network \( g_{\mathrm{proj}}(\cdot) \) that maps \(\mathbf{e}_{\mathrm{3D}}\) into a semantically rich, text-compatible space:
\begin{equation} 
    \mathbf{z}_{\mathrm{3D}} = g_{\mathrm{proj}}(\mathbf{e}_{\mathrm{3D}}),
\end{equation}
where \( \mathbf{z}_{\mathrm{3D}} \in \mathbb{R}^{d'} \). The design follows~\cite{qi2024shapellm}, incorporating three MLP projectors: a local feature projector enhancing fine-grained geometric details, a global feature projector preserving overall semantics, and an absolute position encoding projector injecting spatial cues.

During training, textual embeddings \( \mathbf{z}_{\mathrm{text}} \) are extracted from a pretrained language model, and cross-modal alignment is enforced via a contrastive loss:
\begin{equation} 
\label{eq:loss_align} 
    \mathcal{L} = \mathcal{L}_{\mathrm{align}}\left(\mathbf{z}_{\mathrm{3D}}, \mathbf{z}_{\mathrm{text}}\right).
\end{equation}
This multi-projector setup enhances alignment by capturing diverse structural aspects, improving robustness in the shared embedding space.

\paragraph{Two-Stage Learning with CoT.}
We adopt a two-stage training framework to progressively refine 3D-text alignment, incorporating both CoT and non-CoT data.

\begin{itemize}[leftmargin=1.5em]
    \item \textbf{Stage 1 (Multimodal Alignment)}:  
    The language model remains frozen, and only the projection module is optimized to align 3D features with textual embeddings. Input descriptions follow three annotation formats:
    \begin{itemize}
        \item \textbf{No CoT}: Concise object descriptions without explicit reasoning.
        \item \textbf{Unmarked CoT}: Multi-step reasoning presented as continuous text.
        \item \textbf{Tagged CoT}: Reasoning explicitly segmented with \texttt{<think>} markers.
    \end{itemize}
    
    \item \textbf{Stage 2 (Cross-Modal Adaptation)}:  
    The language model is partially or fully unfrozen to further integrate multimodal reasoning. This phase continues training on both CoT and No CoT samples, ensuring the model retains generalization capabilities while benefiting from structured reasoning. We compare:
    \begin{itemize}
        \item \textbf{Unmarked vs. Tagged CoT}: Assessing whether explicit segmentation improves structured reasoning.
        \item \textbf{No CoT vs. CoT}: Evaluating if stepwise textual reasoning enhances affordance and causal inference.
    \end{itemize}
\end{itemize}

\paragraph{Generality and Extensions.}
Our approach, though focused on point-cloud data, generalizes to other modalities (e.g., 2D images). Recent studies even model point clouds as structured images~\cite{kang2024point}, highlighting the potential for cross-modal adaptation. The key insight is that CoT-structured text provides explicit intermediate reasoning steps, bridging the gap between low-level visual features and high-level semantic understanding.

The two-stage strategy—first aligning 3D representations with text, then refining reasoning integration—is adaptable to various architectures that support projection into a shared embedding space. Additionally, this method can be extended to other domains requiring structured reasoning, such as robotics, medical imaging, or embodied AI.

In summary, our framework systematically integrates chain-of-thought reasoning into 3D vision-language alignment via a robust benchmark, with a structured dataset and a progressive training pipeline. Experimental results (Section~\ref{sec:experiments}) confirm its effectiveness in improving functional and causal inference, establishing a scalable paradigm for multimodal understanding.

%% file: sec/4_experiments.tex
\section{Experiments}
\label{sec:experiments}

We evaluate the impact of CoT annotations on 3D vision-language alignment through a two-stage training strategy and systematic benchmarking on the proposed \textit{3D-CoT Benchmark}. Our analysis examines model performance across different reasoning structures, compares LRMs and LLMs in leveraging CoT, and explores the influence of annotation styles. We further assess qualitative outputs and discuss limitations to provide deeper insights into CoT’s role in multimodal alignment.

\subsection{Implementation Details}
\label{subsec:exp_setup}

\paragraph{Models.}
We compare two categories of language models (LMs) that differ in their innate reasoning capabilities:
\begin{itemize}[leftmargin=1em]
    \item \textbf{LRM}: \textit{DeepSeek-R1-Distill-Llama-8B}~\cite{guo2025deepseek}, which is specifically designed for multi-step inference. Its distilled reasoning module aims to segment complex linguistic inputs into structured steps, potentially leveraging chain-of-thought cues more effectively.
    \item \textbf{LLM}: \textit{Llama-3.1-8B-Instruct}~\cite{grattafiori2024llama}, an instruction-tuned model capable of handling diverse textual prompts but lacking a built-in multi-step reasoning mechanism.
\end{itemize}
Both are integrated into our 3D encoder + projection framework (Section~\ref{sec:method}) to learn to decode/reason shared 3D-text embedding. 

\paragraph{Datasets.}
We conduct experiments on the proposed 3D-CoT Benchmark (Section~\ref{subsec:3dcot_benchmark}), which includes structured annotations for shape recognition, functional inference, and interaction-based reasoning. Each subset is split into 80\% training, 10\% validation, and 10\% test sets, ensuring that no shape instances overlap across partitions.
For evaluation, we use two subsets:
\begin{itemize}[leftmargin=1em]
    \item \textbf{CoT-CAP3D}: Focuses on object affordances and category-level reasoning.
    \item \textbf{CoT-GApartNet}: Covers part-level functional interactions and causal relationships.
\end{itemize}
Each subset is evaluated independently, and we also assess cross-subset generalization, testing models trained on one subset against the other.

\paragraph{Training Protocol.}
We employ a two-stage training strategy to optimize model alignment and reasoning capabilities. 

In Stage 1, we freeze all language model parameters and train only the projection module to align 3D features with text embeddings. This phase uses the AdamW optimizer with a learning rate of 2e-3, a batch size of 256, and a single epoch per dataset. Training follows a linear warm-up for the first 3\% of total steps, followed by cosine decay, with gradient clipping at 1.0 for stability.

In Stage 2, we unfreeze selected layers of the language model to integrate multi-step reasoning signals. This fine-tuning stage employs a reduced learning rate of 2e-5 and a batch size of 128, maintaining the same warm-up and decay schedule as Stage 1.

All experiments are conducted on 4× NVIDIA A800 GPUs, with an average training duration of 24 GPU-hours across both stages. The key hyperparameters for each stage are summarized in Table~\ref{tab:hyperparams}.

\begin{table}[!t]
    \centering
    \captionsetup{justification=raggedright, singlelinecheck=false} 
    \resizebox{0.95\linewidth}{!}{ 
    \begin{tabular}{@{}lcc@{}}  
        \toprule
        \textbf{Hyperparameter} & \textbf{Stage 1} & \textbf{Stage 2} \\  
        \midrule
        Optimizer & AdamW & AdamW \\
        Learning Rate & 2e-3 & 2e-5 \\
        Batch Size & 256 & 128 \\
        Epochs & 1 & 1 \\
        Warm-up Ratio & 3\% of total steps & 3\% of total steps \\
        Gradient Clipping & 1.0 & 1.0 \\
        \bottomrule
    \end{tabular}}
    \vspace{-7pt}
    \caption{Statistics of the training hyperparameters.}
    \vspace{-7pt}
    \label{tab:hyperparams}
\end{table}


\paragraph{Evaluation Metrics.}
To rigorously evaluate the multimodal reasoning capabilities of our alignment approach, we propose a dual-layered evaluation framework that separately quantifies intermediate reasoning quality and the correctness of the model’s final conclusions.

\begin{table*}[t!]
    \centering
    \renewcommand{\arraystretch}{1.1}  
    \setlength{\tabcolsep}{6pt}        
    \resizebox{\textwidth}{!}{         
    \begin{tabular}{llcccccc}  
    \toprule
    \multicolumn{2}{c}{\textbf{Model}} & \textbf{Annotation} & \textbf{TRU} $\uparrow$ & \textbf{COMP} $\uparrow$ & \textbf{OBJ} $\uparrow$ & \textbf{FUNC} $\uparrow$ & \textbf{INTER} $\uparrow$ \\
    \midrule
    \multirow{3}{*}{\textbf{LLM}}  
     & Llama-3.1-8B-Instruct   & No CoT        & \textbf{3.18} $\pm$ 1.42 & 3.30 $\pm$ 1.40 & \textemdash{}  & \textemdash{}  & \textemdash{}  \\ 
     & Llama-3.1-8B-Instruct   & Unmarked CoT  & 3.05 $\pm$ 0.94 & 3.35 $\pm$ 1.11 & 2.85 $\pm$ 1.11 & 2.91 $\pm$ 0.91 & 2.88 $\pm$ 1.48 \\ 
     & Llama-3.1-8B-Instruct   & Tagged CoT    & 2.83 $\pm$ 1.45 & 2.89 $\pm$ 1.42 & 3.18 $\pm$ 1.17 & 2.94 $\pm$ 1.05 & 2.29 $\pm$ 0.93 \\ 
    \midrule
    \multirow{3}{*}{\textbf{LRM}}  
     & DeepSeek-R1-Distill-Llama-8B  & No CoT        & 3.10 $\pm$ 1.30 & \textbf{3.40} $\pm$ 1.24 & \textemdash{}  & \textemdash{}  & \textemdash{}  \\ 
     & DeepSeek-R1-Distill-Llama-8B  & Unmarked CoT  & 2.98 $\pm$ 1.41 & 3.39 $\pm$ 1.38 & \textbf{3.26} $\pm$ 1.26 & \textbf{3.28} $\pm$ 1.12 & \textbf{3.21} $\pm$ 1.19 \\ 
     & DeepSeek-R1-Distill-Llama-8B  & Tagged CoT    & 2.78 $\pm$ 1.39 & 3.09 $\pm$ 1.46 & 3.20 $\pm$ 1.29 & 3.18 $\pm$ 1.13 & 3.06 $\pm$ 1.16 \\ 
    \bottomrule
    \end{tabular}
    }
    \vspace{-5pt}
    \caption{Evaluation results on the \textbf{CoT-CAP3D test set}. Models trained with CoT annotations demonstrate improved reasoning capabilities. LRM consistently outperforms LLM, highlighting better utilization of structured reasoning. Within LRM, \textit{Unmarked CoT} surpasses \textit{Tagged CoT}, suggesting interference from explicit \texttt{<think>} markers. Conversely, for LLM, \textit{Tagged CoT} outperforms \textit{Unmarked CoT}, indicating that encapsulated reasoning aids alignment. Scores are reported as mean~$\pm$~standard deviation. Higher is better.}
    \vspace{-5pt}
    \label{tab:main_results_v2}
\end{table*}

Specifically, motivated by the inherently hierarchical nature of CoT reasoning as same as our dataset constructure, we evaluate intermediate reasoning across three progressively complex cognitive dimensions: object recognition (\textbf{OBJ}), functional reasoning (\textbf{FUNC}), and interaction prediction (\textbf{INTER}), each rated on a 1–5 scale. This incremental hierarchy explicitly mirrors how the CoT-structured data decomposes complex 3D understanding tasks into progressively deeper cognitive steps—beginning from visual identification, moving through inference of object utility, and culminating in logical predictions of plausible human–object interactions.

Further, recognizing that intermediate reasoning (CoT) may occasionally contain inaccuracies, we separately assess the final model-generated conclusions using two dedicated metrics: Truthfulness (\textbf{TRU}) and Completeness (\textbf{COMP}). TRU captures the factual alignment between the model’s final conclusion and the ground-truth reference, ensuring robustness even if intermediate reasoning contains inaccuracies. COMP, on the other hand, evaluates the internal coherence between the model’s CoT reasoning and its final conclusion, verifying whether essential reasoning steps have been faithfully preserved without oversimplification or distortion.
This carefully designed dual-layer evaluation thus isolates the reliability of model-generated conclusions from potential errors propagated through intermediate reasoning, providing deeper insights into the strengths and limitations of CoT-enhanced multimodal alignment strategies. 

\subsection{Main Results}
\label{subsec:main_results}

Table~\ref{tab:main_results_v2} presents the evaluation results on CoT-CAP3D. We have the following key findings: 
1) \textbf{CoT Improves 3D Reasoning.} Models without CoT annotations struggle with OBJ, FUNC, and INTER tasks, relying on superficial descriptions. In contrast, CoT-structured training strengthens multimodal alignment by enforcing stepwise reasoning.
2) \textbf{LRMs Benefit More from CoT.} LRMs consistently outperform LLMs across all CoT conditions, demonstrating superior integration of structured reasoning for 3D-text alignment.
3) \textbf{Annotation Style Affects Model Performance.} Unmarked CoT aligns better with LRMs, likely due to fewer conflicts with their inherent reasoning. Conversely, LLMs benefit from Tagged CoT, which provides clearer reasoning segmentation.
These results confirm CoT's role in improving multimodal alignment, LRMs’ advantage in structured reasoning, and the impact of annotation style on model performance.

\begin{figure}[t!]
    \centering
    \includegraphics[width=0.92\linewidth]{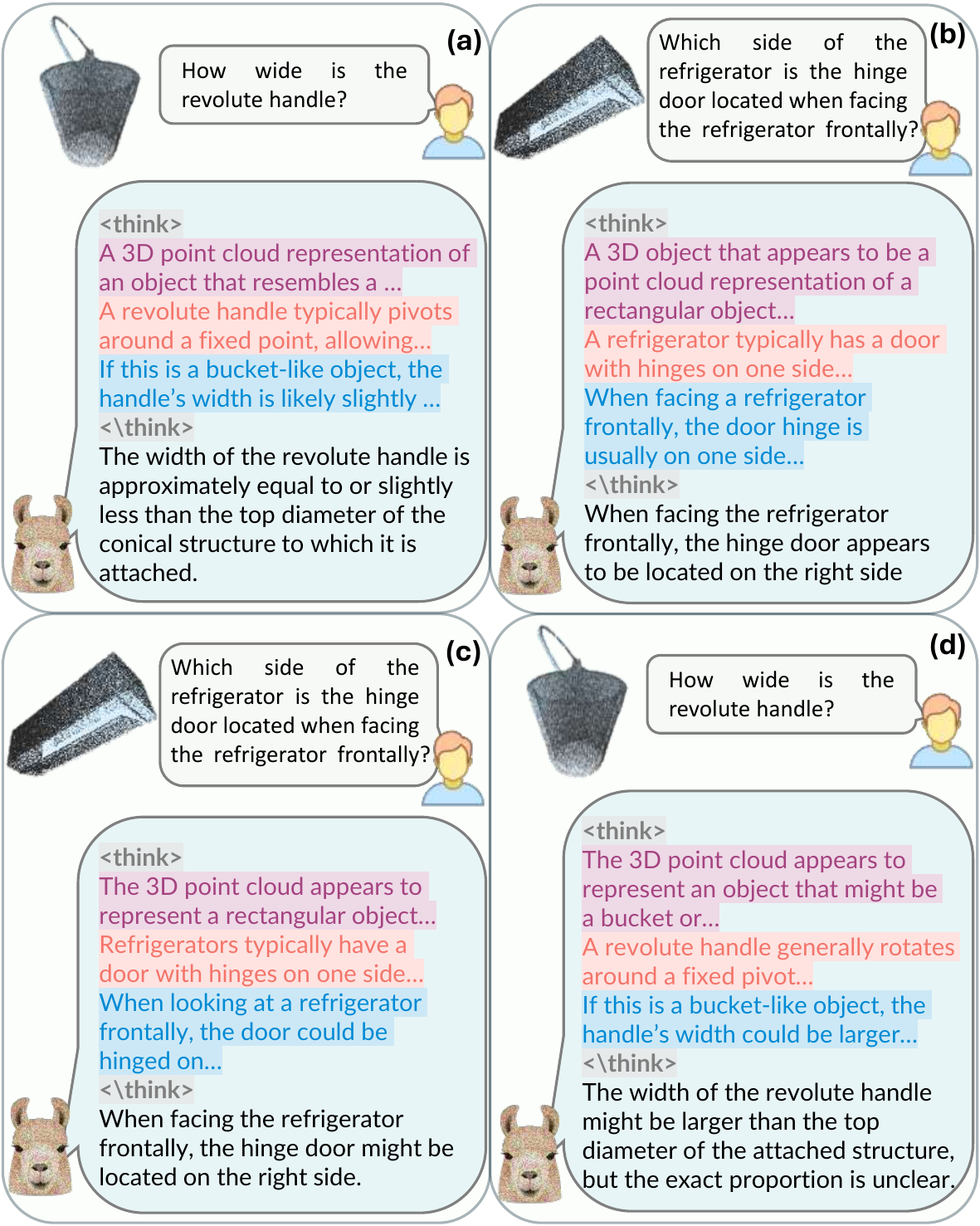}
    \vspace{-0.5em}
    \caption{Example outputs from models evaluated on the \textbf{CoT-GApartNet test set}. The structured annotations facilitate part-level functional reasoning, demonstrating the dataset's effectiveness in enhancing multimodal understanding. a,c from LRM, b,d from LLM.}
    \label{fig:gapartnet_results}
    \vspace{-1em}
\end{figure}


In addition to evaluating CoT-structured reasoning on category-level affordances, we assess model performance on CoT-GApartNet, which focuses on part-level functional interactions. This experiment serves as a complementary validation of our benchmark, demonstrating the effectiveness of CoT-GApartNet annotations in training models for more fine-grained reasoning tasks.
Models trained on CoT-GApartNet achieve consistently strong results on its corresponding evaluation set, particularly in INTER. This indicates that \textbf{CoT-structured annotations effectively support part-level functional understanding}. While CoT-CAP3D-trained models exhibit performance drops when applied to CoT-GApartNet, the reverse is not as pronounced, suggesting that part-based affordance knowledge contributes to a broader understanding of object functionality. 
Figure~\ref{fig:gapartnet_results} provides representative model outputs on CoT-GApartNet tasks. The results highlight how CoT-enhanced reasoning improves the ability to infer mechanical constraints and functional dependencies at the part level.

\subsection{Qualitative Analysis}
\label{subsec:qualitative}


To complement our quantitative findings, we conduct a qualitative analysis of sample model outputs. Figure~\ref{fig:case_study_updated} illustrates a representative example. Our analysis reveals that models trained without CoT tend to produce brief but superficial descriptions, often missing critical affordances or interaction details. In contrast, models utilizing unmarked CoT demonstrate stepwise reasoning, seamlessly integrating geometric and functional insights into a coherent flow. Tagged CoT further enhances interpretability by explicitly segmenting the reasoning process; however, this approach occasionally compromises coherence, particularly in LRM models, where the segmentation can disrupt the natural flow of reasoning. These observations highlight the nuanced trade-offs between interpretability and coherence in CoT-enhanced models.

\subsection{Discussions}
\label{subsec:discussion}

Based on the experimental results, we further discuss the following aspects:  
\begin{itemize}
    \item \textbf{CoT as a Structural Bridge for 3D Reasoning.}  CoT annotations enhance multimodal alignment by explicitly modeling hierarchical reasoning, enabling models to bridge low-level geometry with functional and causal inferences. Without CoT, models rely on surface-level descriptions, struggling with affordance recognition and interaction prediction.  
    \item \textbf{Model-Specific Adaptation to CoT.}  LRM benefits more from CoT due to its inherent multi-step inference capabilities, while LLM gains less but still improves when CoT is structured within explicit markers. These differences suggest that leveraging CoT effectively requires adapting annotation structures to a model’s internal reasoning paradigm.  
    \item \textbf{Implicit vs. Explicit CoT Representation.} Unmarked CoT aligns better with LRM’s inference mechanisms, whereas LLM requires explicit segmentation for structured reasoning. This contrast indicates that models with pre-existing multi-step reasoning may not require explicit CoT markers, while others rely on them for structured alignment.  
    \item \textbf{Challenges in Generalization.}  Although CoT improves cross-task generalization, its benefits are dataset-dependent. Models trained on CoT-CAP3D transfer some knowledge to CoT-GApartNet but do not generalize fully, suggesting that further refinements, such as adaptive CoT formulations, are needed to enhance transferability across diverse 3D reasoning tasks.
\end{itemize}

%% file: sec/5_conclusion.tex
\section{Conclusion}
\label{sec:conclusion}

\begin{figure}[t!]
    \centering
    \includegraphics[width=0.92\linewidth]{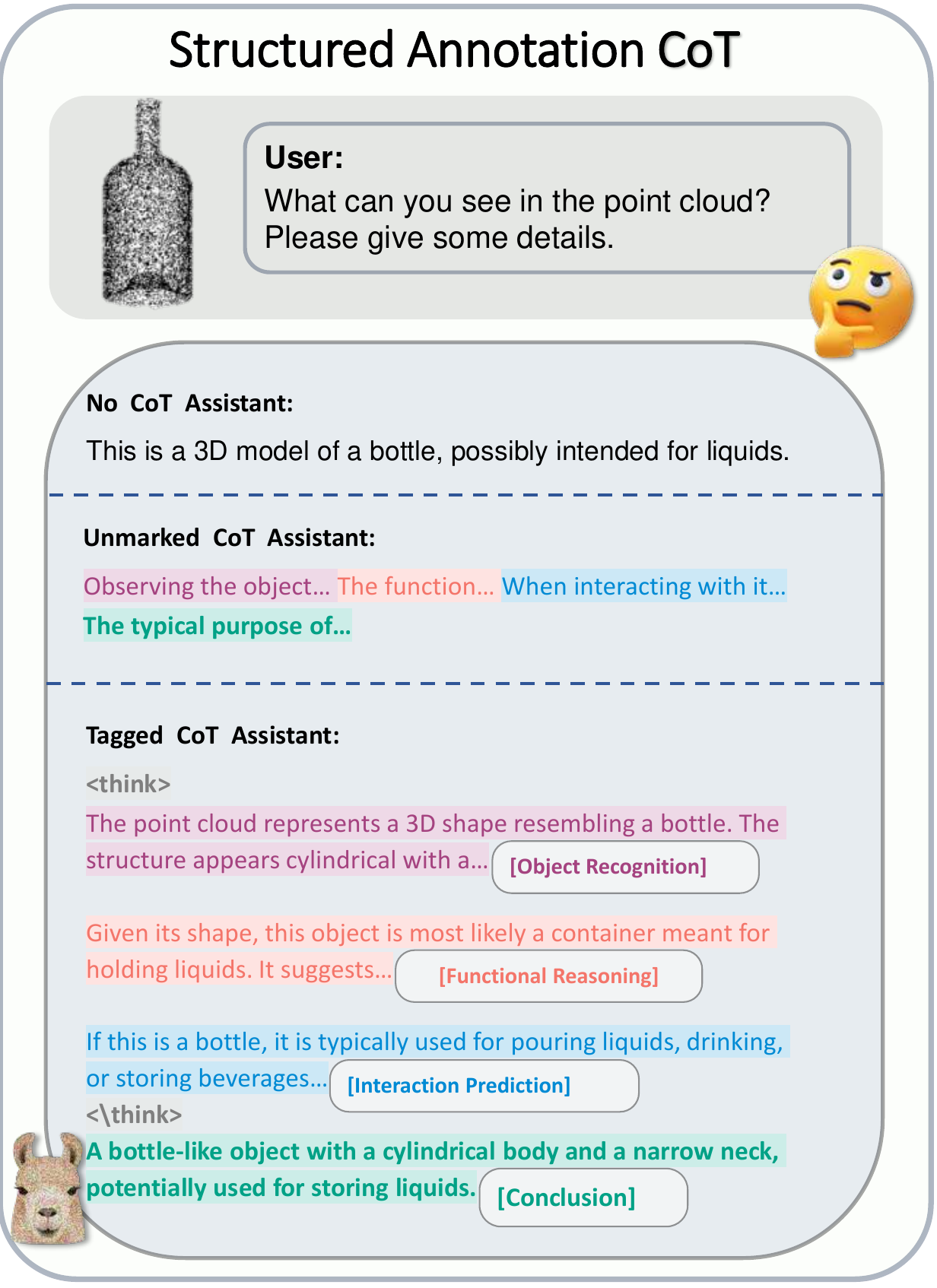}
    \vspace{-0.5em}
    \caption{Case study of model outputs with different annotation strategies. \textit{No CoT} results in minimal descriptions, while \textit{unmarked CoT} and \textit{tagged CoT} produce structured multi-step reasoning. Here, the tagged CoT explicitly segments reasoning with \texttt{<think>} markers.}
    \label{fig:case_study_updated}
    \vspace{-1em}
\end{figure}

Integrating CoT reasoning into 3D vision-language alignment improves semantic grounding, functional inference, and interaction prediction, as demonstrated through the \textit{3D-CoT Benchmark}. LRMs, optimized for multi-step reasoning, leverage CoT more effectively than LLMs, but annotation structure influences performance—unmarked CoT aligns better with LRM inference patterns, likely due to conflicts between CoT annotations and its intrinsic reasoning mechanisms, while explicit reasoning markers enhance LLM performance by enforcing structured segmentation. Despite these benefits, scaling CoT annotations remains resource-intensive, and LRMs incur higher inference costs. Future directions include automating CoT generation, refining annotation strategies tailored to model architectures, and developing efficient reasoning-aware multimodal frameworks.